\documentclass[runningheads]{llncs}
\usepackage{graphicx}
\usepackage{booktabs} 
\usepackage{color}
\usepackage{enumitem}
\begin{document}
\title{What Text Design Characterizes Book Genres?}
%
%
\author{Anonymous}

\author{
Daichi Haraguchi\orcidID{0000-0002-3109-9053}\and
Brian Kenji Iwana\orcidID{0000-0002-5146-6818}\and
Seiichi Uchida\orcidID{0000-0001-8592-7566}}
\authorrunning{D. Haraguchi et al.}

\institute{Kyushu University, Fukuoka, Japan\\
\email{\{daichi.haraguchi, brian, seiichi.uchida\}@human.ait.kyushu-u.ac.jp}}
%
%
\maketitle              
\begin{abstract}
This study analyzes the relationship between non-verbal information (e.g., genres) and text design (e.g., font style, character color, etc.) through the classification of book genres using text design on book covers.
Text images have both semantic information about the word itself and other information (non-semantic information or visual design), such as font style, character color, etc. 
When we read a word printed on some materials, we receive impressions or other information from both the word itself and the visual design.
Basically, we can understand verbal information only from semantic information, i.e., the words themselves; however, we can consider that text design is helpful for understanding other additional information (i.e., non-verbal information), such as impressions, genre, etc.
To investigate the effect of text design, we analyze text design using words printed on book covers and their genres in two scenarios.
First, we attempted to understand the importance of visual design for determining the genre (i.e., non-verbal information) of books by analyzing the differences in the relationship between semantic information/visual design and genres.
In the experiment, we found that semantic information is sufficient to determine the genre; however, text design is helpful in adding more discriminative features for book genres.
Second, we investigated the effect of each text design on book genres.
As a result, we found that each text design characterizes some book genres.  
For example, font style is useful to add more discriminative features for genres of ``Mystery, Thriller \& Suspense'' and ``Christian books \& Bibles.''

\end{abstract}
\section{Introduction}
\begin{figure}[t]
  \centering
  \includegraphics[width=\textwidth]{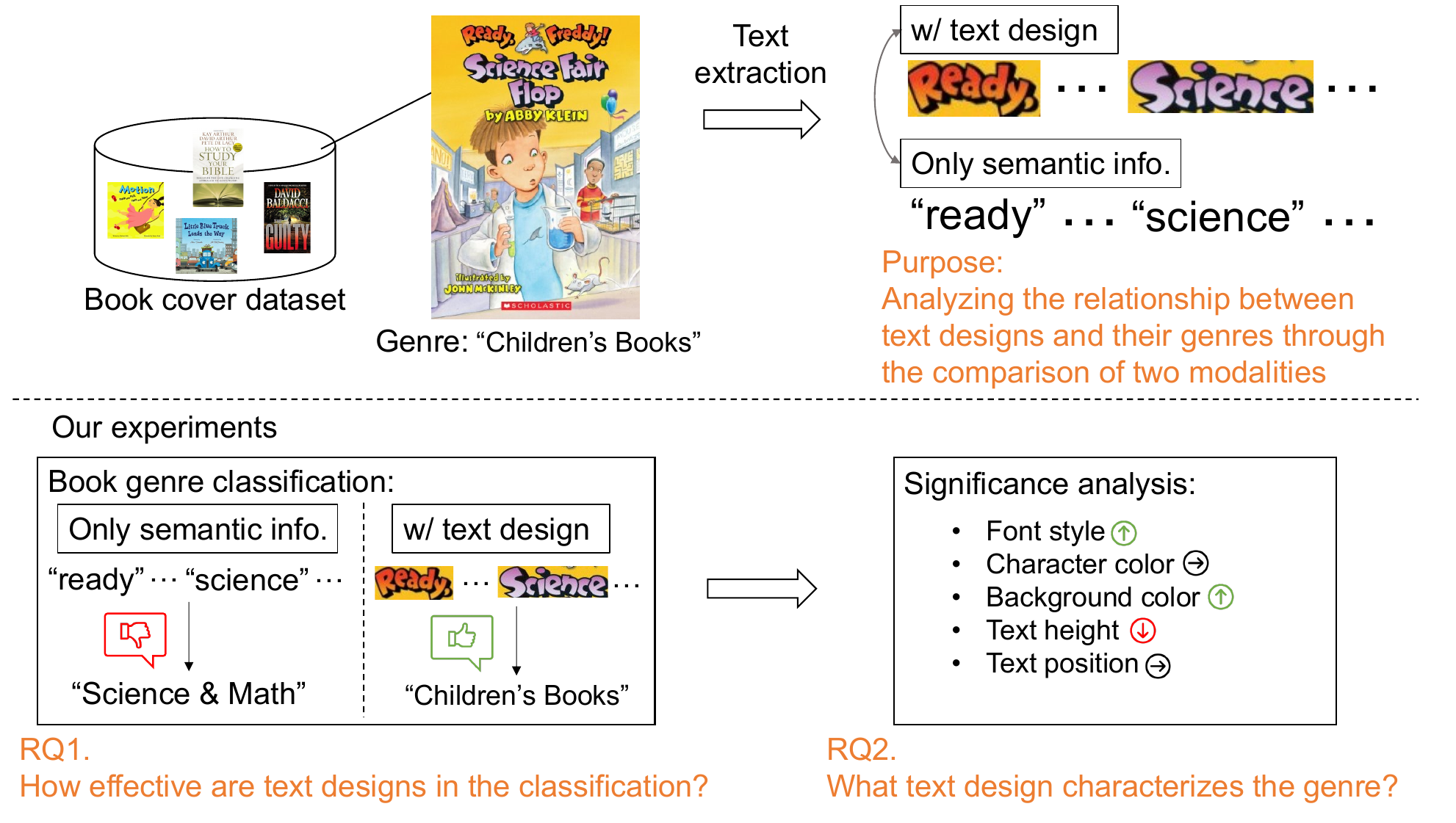}\\[-3mm]
  \caption{Overview of our experiments. We analyze text design on book covers in two scenarios. First, we compare the effectiveness of text design for book genres to semantic information. Second, we analyze what text design characterizes the book genres. }
  \label{fig:teaser}
\end{figure}
Text design plays a crucial role in conveying non-verbal information, such as giving an impression of the media. 
For example, as shown in Figure~\ref{fig:teaser}, a font using a childish fancy style gives an informal and casual impression. 
 
Furthermore, text design is not only font style but also character color, background color, font size, position, etc.
For example, Ikoma, \textit{et al.}~\cite{ikoma2020effect} reported that ``lemon'' and ``strawberry'' are often printed in yellow and red on book covers, respectively.
These colors fit these two fruits colors.
It can be considered that various text design factors are carefully selected by designers to convey non-verbal information, such as impression and genre, etc. 

This paper aims to analyze the relationship between text designs and their non-verbal information (i.e., impression or genre) using book covers.
We use book covers for the following two reasons.
First, we can use genres as non-verbal information.
Second, book covers have a correlation between book genres and their text designs~\cite{yasukochi2023analyzing,shinahara2019serif}. 
Due to being important as the first interaction between a book and the reader, book covers are carefully designed (i.e., design choices such as text are carefully chosen) by designers.

In this paper, we analyze the relationship between text design on book covers and the genres through the classification of book genres 
from two viewpoints.
First, we confirm that the text design contributes to the genre classification task.
To do this, we incorporate semantic features and text design features in a classifier.
By comparing the use of text design features to only semantic features, we are able to elucidate the importance of the text design elements.

Second, we analyze what text designs characterize the genres.
We do this by examining the effects of each design feature when removed from the classifier as well as examining the attention given to each feature. 
For example, in our experiments, we find that removing the font style leads to misclassifications of the genre ``Mystery, Thriller \& Suspense''.
This infers that font style is important to discriminate ``Mystery, Thriller \& Suspense''.

We summarize the two viewpoints as two research questions.
\begin{itemize}[left=17pt]
    \item[RQ1.] How effective is text design in determining book genres compared to semantic information?
    \item[RQ2.] What text design characterizes book genres?
\end{itemize}

\par 
Obtaining these answers to the above questions might be helpful for non-expert people to understand designers' knowledge.
Recently, some studies have tried to generate text images using generative models~\cite{wang2022aesthetic,miyazono2021font}. 
Introducing designers' knowledge into such models leads to improving text design to be more effective for the context or situation.

Our contributions are as follows.
\begin{itemize}
    \item To the best of the authors' knowledge, this is the first attempt to quantitatively evaluate that text designs add more discriminative features for book genres.
    \item Our experiment shows that semantic features are sufficient to determine book genres; however, text design contributes to boosting the classification results. 
    \item In our detailed analysis of text design in the classification task, we found what text design is and how it is important. For example, font style is useful to add more discriminative features for the genre of ``Mystery, Thriller \& Suspense,'' and character color is useful for the genre of ``Romance.''
\end{itemize}
\section{Related Work}
\subsection{Text Design Analysis}
In the marketing field, there are many studies about text design, especially font style, for products, logos, and advertisements.
Henderson,~\textit{et al.}~\cite{henderson2004impression} created guidelines to help effectively select typefaces.
For creating the guidelines, they analyzed the important impression coming from the typeface and the characteristics explaining the typeface.
Doyle,~\textit{et al.}~\cite{doyle2006dressed} measured the connotative meaning included in fonts and product categories and analyzed how these factors are combined.
Liao,~\textit{et al.}~\cite{liao2015emotional} analyzed the emotional responses of consumers to food packaging. They especially addressed three typical design elements of food packaging, including font style. 

Other text design analyses have also been conducted.
Some studies analyze text design, such as font style and text color on book covers~\cite{ikoma2020effect,shinahara2019serif}.
Kulahcioglu,~\textit{et al.}~\cite{kulahcioglu2019paralinguistic} investigated font style and text color that matches the word clouds.
Shirani,~\textit{et al.}~\cite{shirani2020let} associated the visual attributes of fonts with the verbal context.

\subsection{Classification of Book Genres}
A classification of book genres has been attempted by using various inputs.
Iwana,~\textit{et al.}~\cite{iwana2016judging} are pioneers of the classification of book genres. They proposed a book cover dataset.
Following their study, many studies have tried the classification of book genres using different inputs.
Biradar,~\textit{et al.}~\cite{biradar2019classification} and Lucieri,~\textit{et al.}~\cite{lucieri2020benchmarking} used book covers and their title as input for the classification.
Kundu,~\textit{et al.}~\cite{kundu2020deep} employed book covers and detected text from book covers for the classification.
Worsham,~\textit{et al.}~\cite{worsham2018genre} attempted to the classification the book genres using the texts included in the corpus of literature.

A lot of studies have tried the classification using book covers and text printed on book covers.
However, there is no study that analyzes text designs that print on book covers through book genre classification.
In this paper, we attempt to classification of book genres using text designs.
Through the classification, we analyze the relationship between text design on book covers and their genres. 

\section{Dataset and Feature extraction}\label{sec:dataset}
\subsection{Book Covers}
We used the book cover dataset proposed by Iwana,~\textit{et al.}~\cite{iwana2016judging} in our experiments.
The reason for using book covers to analyze text design is that book covers are carefully designed by experts, and appropriate text designs are selected.
Therefore, it is expected that there is a certain relationship between these genres and text designs. 
For this purpose, extracting texts on the book cover images is necessary.
The details of text extraction are described in the next section.

The book cover dataset contains two types of data for classification (Task 1) and data mining (Task 2). We used only Task 1 data in our experiments.
Task 1 data has 30 genres with balanced classes. 
Each genre, i.e. class, contains 1,710 training data and 190 test data.
Note that we used slightly fewer book covers than the original ones because some book cover images do not have text. 

\subsection{Text Images on Book Covers}\label{sec:text_img}
To analyze text design, we collected text images on book covers by text detection and recognition.
We detected words on book covers by using CRAFT~\cite{baek2019character}.
After that, we recognized each word image using TPS-ResNet-BiLSTM-Attn, which is the best text recognition model in~\cite{baek2019wrong}.

We only used the words included in the Google News corpus because we employed word embeddings by word2vec trained by the Google News corpus as semantic features.
Therefore, words that were not used for training word2vec (special proper nouns, coined words, etc.) cannot be used. 
Additionally, we used word images with more sizes than 14 $\times$ 14 to properly extract text design features from word images.
As a result, we obtained 14.6 words per book in the training data.
Note that the average number of words in the genre of ``Test preparation'' is more than 30.
To eliminate the bias effect of the number of words, we set the maximum number of words to 16.

\subsection{Semantic Feature}\label{sec:features}
To extract semantic features from words on book covers, we employed word embeddings by word2vec~\cite{mikolov2013distributed} using the Google News corpus.
Note that all letters were converted to lowercase when obtaining the word embeddings.
The semantic features are 300-dimensional vectors.
To arrange the dimension, we also set the number of dimensions of the other features to 300.

\subsection{Text Design Features} 
\subsubsection{Font style feature} 
We extracted font style features on word images by using ResNet-50~\cite{he2016deep} following other methods~\cite{choi2019assist,kulahcioglu2020fonts,matsumura2020font}.
To extract 300-dimensional vectors from ResNet50, we inserted a fully connected (FC) layer with 300-dimensional output before the last FC layer in ResNet-50.
To train the ResNet-50, we used synthesized font images using Google Fonts~\footnote{\url{https://github.com/google/fonts}} and SynthTIGER~\cite{yim2021synthtiger}, which can synthesize a font into a background image.
In more detail, we used 2,094 fonts from Google Fonts and synthesized 1,000 images by each font. 
We trained the ResNet-50 using 900 images by each font as training and 100 images by each font for validation.
We resized each image's height to 64 or width to 128 with keeping the aspect ratio, and then we added padding to each image to become an aspect ratio of 1:2 (height:width).

\subsubsection{Character color feature}
We extracted the character regions by a similar method of \cite{ikoma2020effect}.
We first binarized the word images by using the Otsu method. Then, we considered the pixels around the bounding box as the background and the others as characters.
We extracted RGB values from character regions.
Then, we counted the number of each value.
We considered the top 100 RGB values as character color features.

\subsubsection{Background color feature} 
As the same method with character color features, we extracted background color features.
We first extracted background regions and RGB values from the background regions.
Then, we counted the number of each value.
We considered the top 100 RGB values as background color features.

\subsubsection{Text height feature} 
We used the height of the bounding box as text height.
To consider relative text height, we divided each height by each height of the book cover. 
To obtain 300-dimensional features, we added an FC layer before the proposed model of the Hierarchical Transformer. 
Please see Section~\ref{sec:model} for the details of the model.
Through the fully connected layer, text heights become 300-dimensional features.
This FC layer is trained simultaneously while training the Hierarchical Transformer.

\subsubsection{Text position feature} 
We used the coordinate of the left top bounding box as the text position.
To consider the relative position, we divided each coordinate by each height and width of the book cover. 
To expand the 2-dimensional text positions to 300-dimensional features, we employed an FC layer before the Hierarchical Transformer.
This FC layer is also trained simultaneously while training the Hierarchical Transformer.

\section{Methodology}

\begin{figure}[t]
    \centering
    \includegraphics[width=0.85\textwidth]{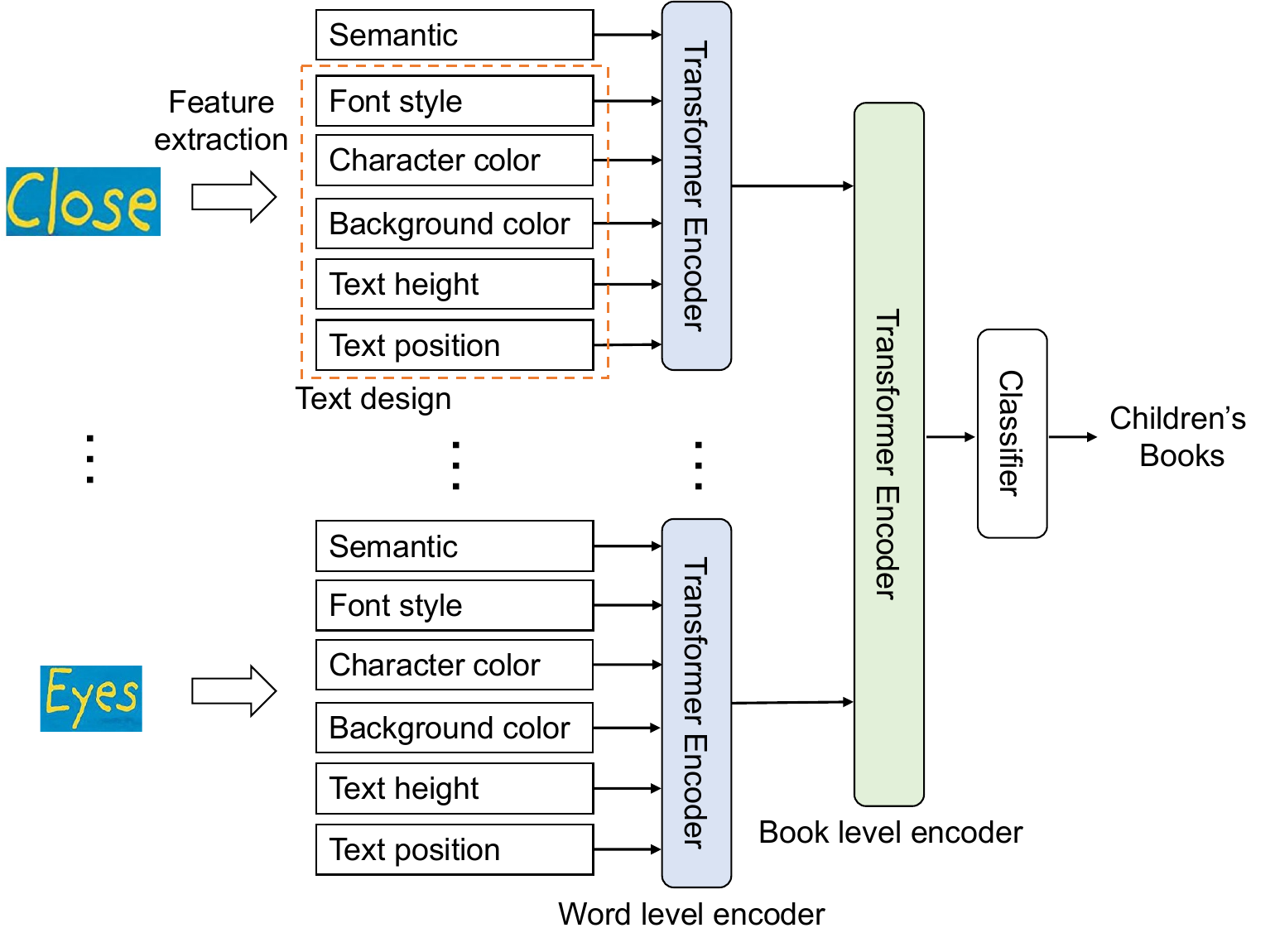}\\[-4mm]
    \caption{The architecture of the Hierarchical Transformer.}
    \label{fig:model}
\end{figure}
\subsection{Overview}

We conducted a classification of book genres using words and text designs on book covers.
We employed font style, character color, background color, text height, and text position as text design.
For the classification, we employed the Hierarchical Transformer, as shown in Figure~\ref{fig:model} (See Section~\ref{sec:model}).
We extracted semantic features and text design features from text images and input these features into the model.
The model estimates a genre based on these features.

\subsection{Hierarchical Transformer for Classification of Book Covers}\label{sec:model}
To classify book genres, we propose a Hierarchical Transformer, which has two levels of encoders.
The first encoder is a word-level encoder, which can encode the text design features of the text into a feature vector.
The self-attention in this encoder weights the relationships between the design elements in order to conduct the classification task. 
This behavior is similar to how designers consider the relationships between various aspects of the design. For example, using dark text on  light backgrounds.

The other encoder is a book-level encoder.
The book-level encoder encodes all features from the word-level encoder into a feature vector. 
Then, the feature vector is fed into a classifier and classified into one of 30 genres.  
Note that we used class tokens output from both encoders as feature vectors. 

Thanks to the hierarchical architecture, we can analyze the contributed elements for the classification in more detail by using attention rollout~\cite{abnar-zuidema-2020-quantifying}.
We can obtain the two types of attention in the Hierarchical Transformer, each relating to the hierarchial level.
The attention in the word-level encoder indicates the contributed elements per text designs and semantic information.
The attention in the book-level encoder indicates the contributed words, including the text designs.
From these two types of attention, it is easy to analyze what text design of what words contributes to the classification.

Our model does not include positional encoding for two reasons.
First, we randomly extract text designs from the book covers. Second, the input features of text position may implicitly include such information.


\subsection{Implementation Detail}
All input feature vectors to the word-level encoder are 300 dimensions as described in Section~\ref{sec:features}.
The maximum number of words is 16 for input, as mentioned in Section~\ref{sec:text_img}.
The number of layers is one and four for the word-level encoder and the book-level encoder, respectively.
The number of heads is six in both encoders.
Feature vectors output from both transformer encoders are 300 dimensions.
The classifier consists of two fully connected layers.
We set the batch size to 64 and used Adam optimizer.
Additionally, we used cross-entropy as a loss function and set the learning rate to $10^{-5}$.

\section{Classification of Book Genres}\label{sec:quantitative}
\begin{table}[t]
\centering
  \caption{Top-$N$ accuracy of classification of book genres (\%).}
  \vspace{-3mm}
  \label{tab:acc}
  \begin{tabular}{lcc}
    \toprule
    Model & Top-1 & Top-3 \\
    \midrule
    Full model (all text design feat.) & \textbf{48.45} & \textbf{68.90}\\
    \hspace{3mm} w/o font style & 47.20 & 67.87\\
    \hspace{3mm} w/o character color & 47.16 & 68.41\\
    \hspace{3mm}  w/o background color& 47.33 & 68.77\\
    \hspace{3mm} w/o character \& background colors & 47.22 & 67.96\\
    \hspace{3mm} w/o text height & 48.25 & \textbf{68.90}\\
    \hspace{3mm} w/o text position & 47.09 & 68.05\\
    \hspace{3mm} w/o text height \& position & 46.86 & 67.73\\
    \hspace{3mm} w/o semantic & 17.06 & 32.69\\ \hline
    Baseline (only semantic feat.) & 45.46 & 67.00 \\
    Random Chance & 3.33 & 10.00 \\ 
  \bottomrule
\end{tabular}
\end{table}
We conducted the classification of book genres under various conditions.
One is the classification using only semantic features as a baseline.
In this classification, we directly input semantic features to the book-level encoder.
The others are classification using all features and removing several of the features.
For example, we trained the model without font style (i.e., w/o font style) and then evaluated the performance.

We show the results in Table~\ref{tab:acc}.
While the baseline demonstrates a good accuracy, adding any of the text design features is able to supplement the information and increase the accuracy. 
Conversely, w/o semantic has a lower accuracy, but the accuracy is still higher than random chance. This indicates that the text design
does contribute to the classification of the book genres.

Under the condition of removing one text design, the accuracy of w/o text position in the top-1 and the accuracy of w/o font style in the top-3 are the lowest.
From these results, text position and font style might be the most effective designs for the classification of book genres. 

The accuracy of w/o text height is not so much different from the Full model.
However, the accuracy of w/o text height \& position is much lower than only w/o text position or w/o text height.
This means that the combination between text height and position is important for the classification.
The other combinations might also be important for the classification.
We will consider the analysis of such combinations in future work.

\section{Text Design Analysis on Book Covers}~\label{sec:text-design-analysis}
\begin{figure}
    \centering
    \includegraphics[width=1\textwidth]{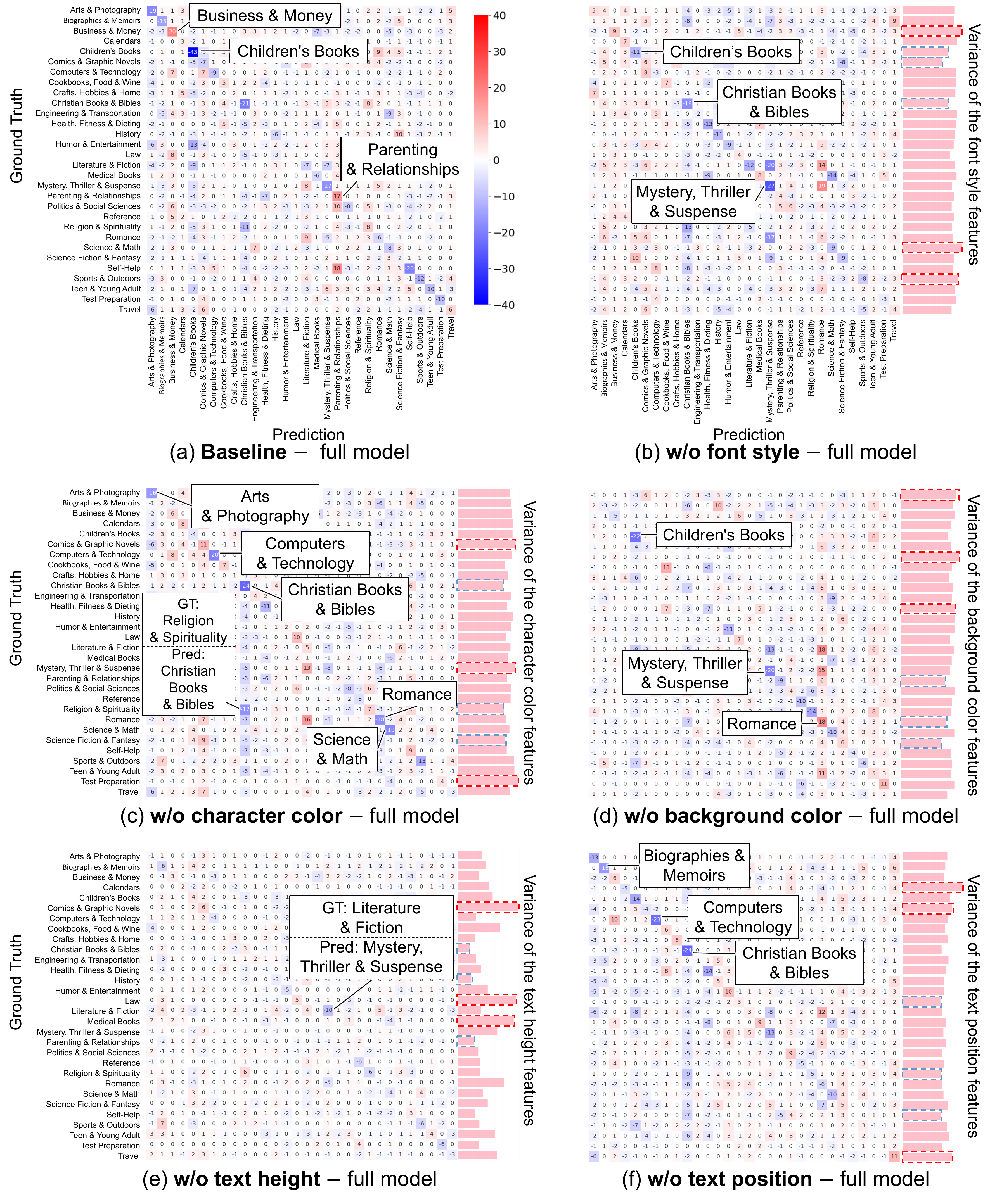}\\[-3mm]
    \caption{The difference between two confusion matrices between the condition of removing features and the full model.
    Each bar plot shows the variance of each text design element. Red dotted boxes show the three highest variances. Blue dotted boxes show the three lowest variances.}
    \label{fig:confusion}
\end{figure}

We analyze the classification result focusing on each text design individually.
To this end, we calculate the difference between the confusion matrix of removing a text design and the confusion matrix of the full model.

Figure~\ref{fig:confusion} shows the difference between two confusion matrices of the classification results.
We summarize the meaning of the color of each cell as follows.
\begin{itemize}
    \item The \color{blue}blue\color{black}~cells on the diagonals show a decrease in incorrectly classified samples by removing the design features. This implies that the design might contribute to classifying the genre.
    \item The \color{red}red\color{black}~cells on the diagonals show an increase in correctly classified samples by removing the design features. This implies that the design might have a negative effect on the classification.
    \item The \color{blue}blue\color{black}~cells not on the diagonals show a decrease in misclassified samples by removing the design features. This means that the design might have a negative effect on the classification.
    \item The \color{red}red\color{black}~cells not on the diagonals show an increase in misclassified samples by removing the design. This means that the design might contribute to classifying the genre.
\end{itemize}
In the following section, we describe the effect of each text design.
Additionally, we summarize the characterization of genres by each text design in Table~\ref{tab:summary}.

\subsection{Font Style}
As shown in Figure~\ref{fig:confusion} (b), ``Christian books \& Bibles'' and ``Mystery, Thriller \& Suspense'' in the diagonal elements are deep blue. From this result, font style contributes to classifying these two genres. 

``Children’s Books'' show blue in the diagonal elements and red in the vertical cells. 
This means that font style contributes to reducing misclassification for ``Children’s Books.'' 

Especially, the variances of font style features of ``Christian books \& Bibles'' and ``Children's Books'' are low. This means that similar fonts tend to be used in each genre. Therefore, the font style contributes to classifying these genres.

Note that the fewer variances do not necessarily contribute to improving the classification result.
For example, the variance of ``Comics \& Graphic Novels'' is low; however, the classification results are not improved.
This is because similar fonts might be used in other genres. 

\subsection{Character Color}
We can see blue cells in the diagonal elements of ``Arts \& Photography,'' ``Computers \& Technology,'' ``Christian Books \& Bibles,'' ``Romance'' and ``Science \& Math.''
For these genres, character color is effective in identifying their genres.

Note that the misclassification of ``Religions \& Spirituality'' to ``Christian Books \& Bibles'' is decreasing by removing character color. This means that these two genres tend to be used in similar colors for their text.
Both genres have low variance. Therefore, this trend might be more remarkable than other genres.

\subsection{Background Color}
We can see blue cells in the diagonal elements of ``Children’s Books'' and ``Mystery, Thriller \& Suspense.''
From this, background colors are effective in classifying these two genres.

For ``Romance,'' it is not effective to classify the genre.
Removing the background color increases the number of correctly classified samples. This means that the color has a negative effect.
Note that the number of misclassified samples as ``Romance'' (see vertical cells) is also increased. Therefore, some samples become discriminative by background color.

\subsection{Height}
The results show that there is no strong effect of text height.
However, the height slightly negatively affects classifying ``Literature \& Fiction.''
For example, by removing the height, the number of samples misclassified as ``Literature \& Fiction'' to ``Mystery, Thriller \& Suspense'' is reduced.
We emphasize that this result did not indicate that text height does not correlate with genres. 
As shown in Table~\ref{tab:acc}, w/o text height \& position is much lower than only w/o text position and w/o text height.
This means that the combination of text height and text position is very important.
It might be more clear trends by analyzing the correlation between the genre and the combination of designs.

\subsection{Position}
Text position has a positive affect on the genres, ``Biographies \& Memoirs,'' ``Computers \& Technology''and ``Christian Books \& Bibles''.
However, ``Christian Books \& Bibles'' also have some misclassified samples (see vertical cells). 
From these results, ``Christian Books \& Bibles'' has a specific trend for the text position; however, a few books of the other genres might have the same style. Therefore, some samples might be misclassified as ``Christian Books \& Bibles''. 

\subsection{All Text Design}
As shown in Figure~\ref{fig:confusion} (a), ``Business \& Money'' and ``Parenting \& Relationship'' in the diagonal elements are red.
From this result, these two genres might have no effect on using text design for the classification.

On the other hand, ``Children’s Books'' in the diagonal elements are deep blue.
This means that text designs are very effective in classifying this genre.
As we mentioned above, ``Children’s Books'' have a strong or slight effect of font style, background color, and other text designs on classifying the genre.

\section{Visualization of Attention}\label{sec:attention}
\begin{figure}[t]
    \centering
    \includegraphics[width=0.9\textwidth]{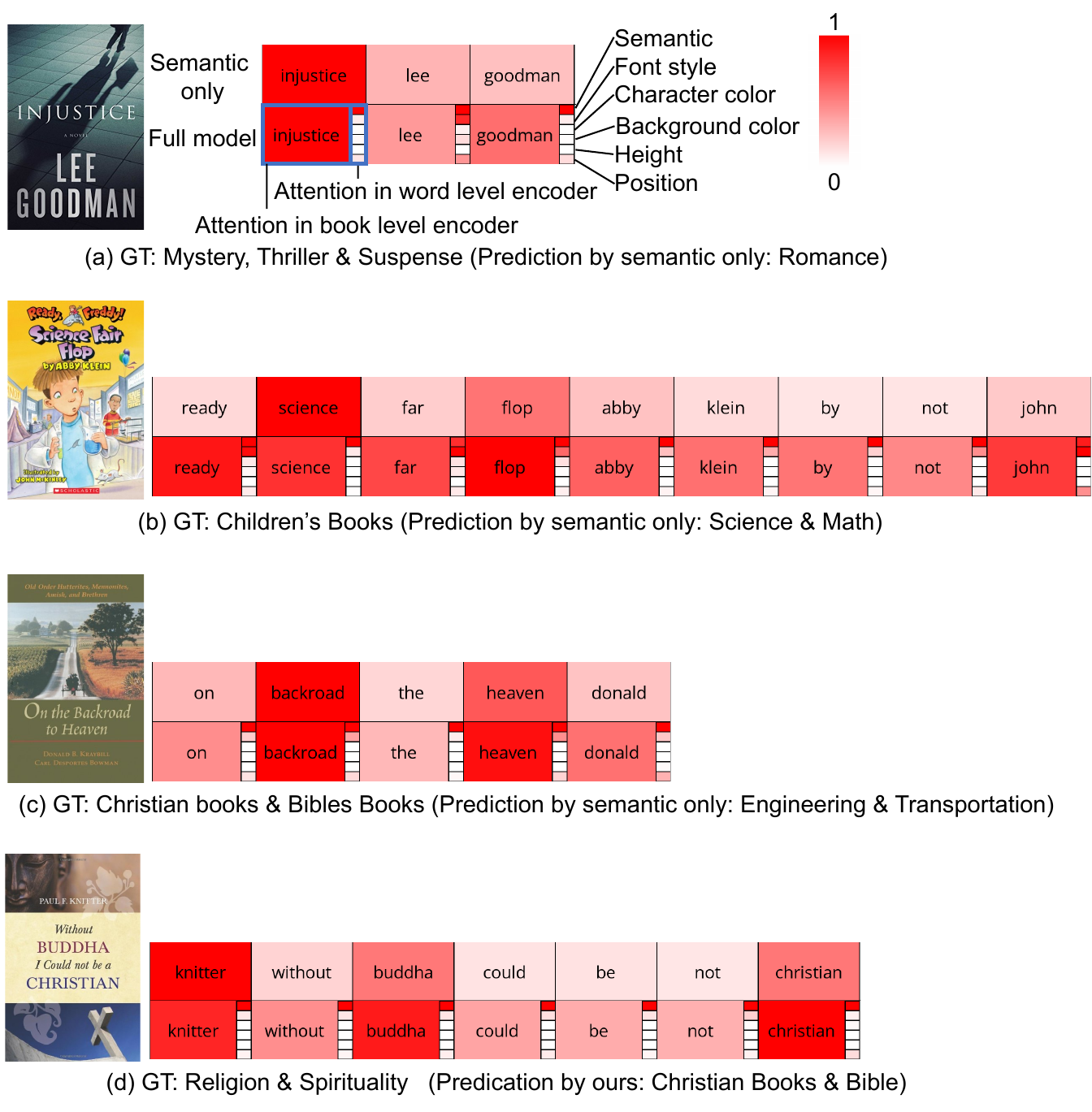}\\[-4mm]
    \caption{Examples of visualization of attention. The top of each subfigure is extracted from the baseline. The bottom of each subfigure is extracted from the full model. The bottom row of each subfigure also shows the attention to the design features. Deeper red shows strong attention. (a) to (c) are correctly predicted samples by ours. (d) is not a wrong sample by ours.}
    \label{fig:attn}
\end{figure}
We visualize the attention in the Hierarchical Transformer by attention rollout~\cite{abnar-zuidema-2020-quantifying}.
Because the proposed Hierarchical Transformer has two levels, the word level and the book level, it is possible to analyze the contribution of the text design for each word. 
Thus, we use this idea in Figure~\ref{fig:attn}, which shows the results of visualization of attention.

Figure~\ref{fig:attn} (a), (b), and (c) show samples that the full model can correctly classify each genre, whereas the baseline (using only semantic features) can not classify them.
Some words in the full model have stronger attention than the baseline.
Such words have strong attention to font style features.
For example, the ``read'' of (b) in the full model has stronger attention than the baseline and shows stronger attention in font style.
Interestingly, the baseline classified the book cover genre as ``Science \& Math'' and had the highest attention on the word ``science.''
On the other hand, the full model could correctly classify it due to considering the text designs.
Additionally, (c) is also the same case.
This sample is also classified as a different genre of ``Engineering \& Transportation'' by the baseline. It might be because the word with the highest attention is ``backroad'' which is related to ``Transportation.''
In contrast, the font styles of ``backroad'' and ``heaven'' have strong attention in the full model, and therefore, the book cover might be correctly classified. 
Additionally, in the full model, some stop words have stronger attention than the baseline (e.g., ``on'' in (c)).
These results indicate that text designs are effective in increasing attention to word level and characterizing the genres.
This leads to an increase in the classification accuracy in some genres.

On the other hand, (d) is a sample that full model misclassified ``Religion \& Spirituality'' to ``Christian books \& Bibles.''
Text designs sometimes negatively affect the classification.
Especially several genres with similar text designs have such trends.

\section{Analysis of Text Design Usage}
\begin{figure}[t]
    \centering
    \includegraphics[width=1\textwidth]{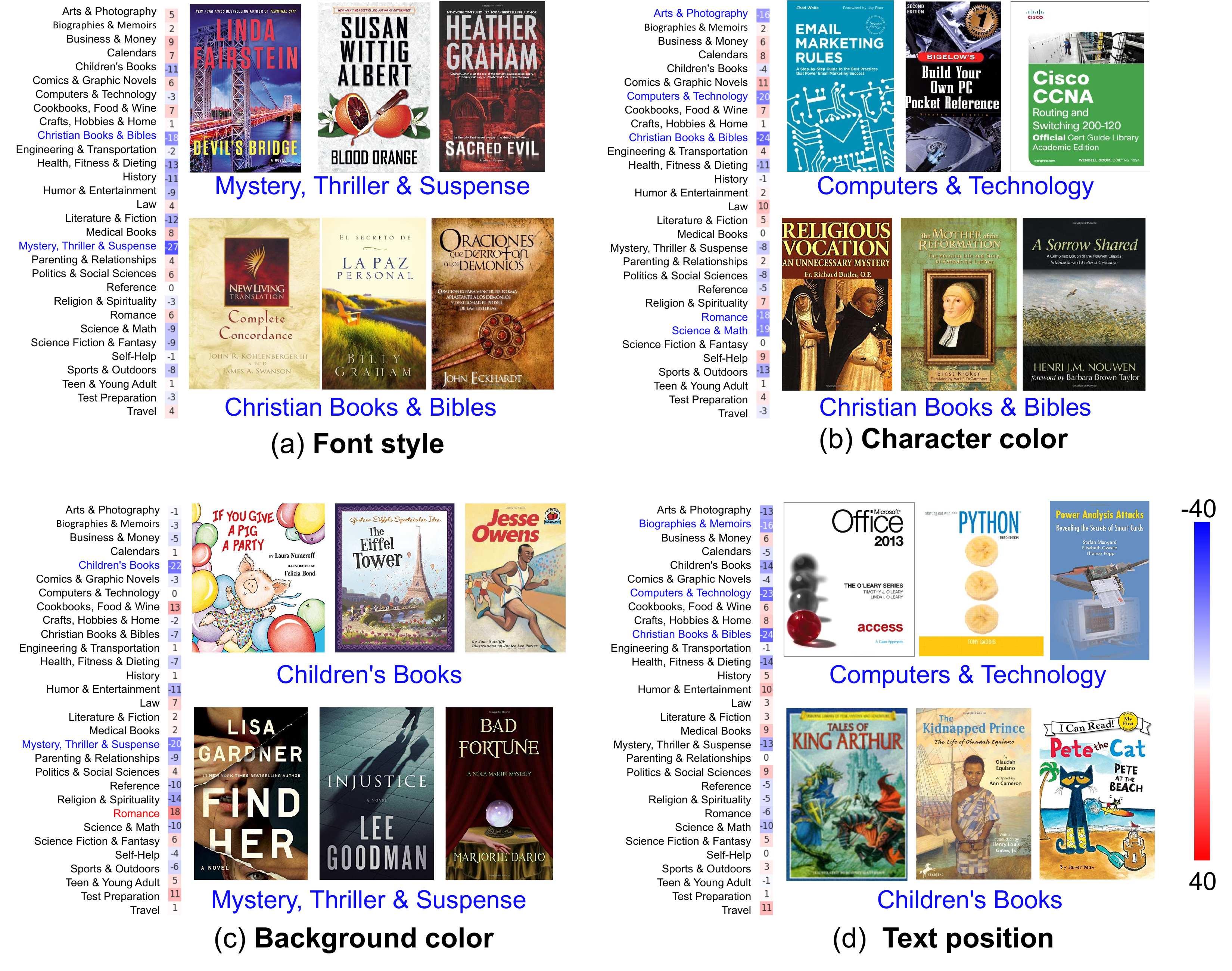}\\[-3mm]
    \caption{Examples of book covers. The values of the left shows the diagonal elements on the confusion matrix shown in Figure~\ref{fig:confusion}.}
    \label{fig:font_example}
\end{figure}
We analyzed the usage of text design in more detail, focusing on text design and book genres, which showed a correlation in Section~\ref{sec:text-design-analysis}.
In other words, we analyzed the samples that showed specific trends in the diagonal cells of each matrix in Figure~\ref{fig:confusion}.

For the analysis, we first conducted k-means clustering by each text design feature on all test data.
We set the $k$ as 5 using the elbow method for each feature, with the exception of font style.
The elbow method and other tests were unable to determine a $k$ for font style, so it was manually set to $k=20$.
Then, we confirmed which cluster the text design of the book cover on the diagonal in Figure~\ref{fig:confusion}.
Finally, we qualitatively evaluated the cluster where the most or second most book covers belonged.

Note that semantic information is more dominant than text designs, as shown in attention analysis in Section~\ref{sec:attention}, and therefore, the contribution of each design is very slight.
Additionally, whether the following elements directly contribute to the classification is ambiguous because the correlation between designs might also be important for the classification.

\subsection{Font style}
``Mystery, Thriller \& Suspense'' used condensed fonts as shown at the top of Figure~\ref{fig:font_example} (a).
``Christian Books \& Bibles'' used fonts with a contrast of stroke thicknesses, such as serif fonts, as shown at the bottom of Figure~\ref{fig:font_example} (a). 
These fonts might not only be serif fonts but also historical script fonts defined in a book~\cite{Seibundo}.  
The historical script fonts might match the traditional atmosphere of the genre ``Christian Books \& Bibles.'' 
Both results correspond to the results of the previous font style analysis~\cite{yasukochi2023analyzing,shinahara2019serif}.

\subsection{Character color}
``Computers \& Technology'' used white text as shown at the top of Figure~\ref{fig:font_example} (b).
``Christian Books \& Bibles'' used yellow or orange in the text as shown at the bottom of Figure~\ref{fig:font_example} (b).
Compared to white in ``Computers \& Technology,'' yellow or orange might not be very common, and therefore, this trend is very interesting.
Though we did not show examples, black texts are used in both genres of book covers in the second most cluster.

\subsection{Background color}
``Children's Books'' used light colors, such as white and light blue, on the background of the text, as shown at the top of Figure~\ref{fig:font_example} (c).
In contrast to the ``Children's Books,'' ``Mystery, Thriller \& Suspense'' used dark colors on the background of the text, as shown at the bottom of Figure~\ref{fig:font_example} (c). This color might match the tense atmosphere of the genre.

\subsection{Text position}
``Computers \& Technology'' and ``Children's Books'' arranged the text on the top of the book covers as shown in Figure~\ref{fig:font_example} (d).
This trend might look like common. However, for example, ``Mystery, Thriller \& Suspense'' shown in Figure~\ref{fig:font_example} (c), several texts are arranged on the bottom.
Considering this fact, the trend of ``Computers \& Technology'' and ``Children's Books'' is important.


\section{Conclusion}
\begin{table}[t]
  \caption{Summary of the characterization of book genres by text designs.}
  \centering
  \vspace{-3mm}
  \label{tab:summary}
  \begin{tabular}{l|p{8 cm}}
    \toprule
    Text design &  Correlated book genres \\ 
    \midrule
    Font style & ``Christian books \& Bibles,'' ``Mystery, Thriller \& Suspense'' and ``Children’s Books'' are correlated with font style usage. In particular,  ``Mystery, Thriller \& Suspense'' used condensed fonts. ``Christian Books \& Bibles'' used fonts with a contrast of stroke thicknesses, such as serif fonts. \\ \hline
    Character color &  ``Arts \& Photography,'' ``Computers \& Technology,'' ``Christian Books \& Bibles,'' ``Romance'' and ``Science \& Math'' are correlated with the character color usage. In particular, ``Computers \& Technology'' used white text, and ``Christian Books \& Bibles'' used yellow or orange text.
Compared to white in ``Computers \& Technology,'' yellow or orange might not be very common, and therefore, this trend is very interesting.\\ \hline
    Background color &  ``Children's Books'' used light colors, such as white and light blue. ``Mystery, Thriller \& Suspense'' used dark colors on the background of the text.  \\ \hline
    Height &  There is no specific use case. However, pay attention to the genre of ``Literature \& Fiction.'' In this genre, text height sometimes has a negative effect.\\ \hline
    Position &  ``Biographies \& Memoirs,'' ``Computers \& Technology''and ``Christian Books \& Bibles'' are correlated with the text position. In particular, ``Computers \& Technology'' and ``Children's Books'' arranged the text on the top of the book covers.\\ 
  \bottomrule
\end{tabular}
\end{table}

This study analyzes the relationship between book genres and text
design through the classification of book genres using text design on book covers.
For text design, we employed font style, character color, background color, text height, and text position.
The classification result showed that semantic information is sufficient to classify book genres; however, text design contributes to slightly increasing the classification accuracy.
The analysis of the classification results showed that each text design characterizes some genres. 

We summarize the answers to the two research questions.
\begin{itemize}[left=17pt]
    \item[RQ1.] How effective is text design in determining book genres compared to semantic information?
    \item[A1.] Text designs are slightly effective in determining the genres, as shown in Table~\ref{tab:acc}. Any text design can add more discriminative features to the text. How effective it is depends on the text design. For example, the font style and text position are the most effective. 

    \item[RQ2.] What text design characterizes book genres?
    \item[A2.] We summarize the text designs and their use case in Tabel~\ref{tab:summary}. 
    For example, if you want to characterize books of ``Mystery, Thriller \& Suspense,'' please pay attention to selecting its font style and background color.
\end{itemize}

In future work, we will conduct a more comprehensive analysis of text design on book covers.
Some text design usage might have a correlation.
For example, a character color and background color might have a correlation because of contrast for readability.
Therefore, we will consider the combination of each text design.
Additionally, we will analyze more detail in each text design using a larger dataset. 
In our detailed analysis of text design usage, we could see several correlations between specific genres and text designs.
For example, ``Mystery, Thriller \& Suspense'' used condensed fonts.
However, the tendencies were very weak.
Therefore, we will prove the tendencies using a larger dataset.
%
%
%
%

\section*{Acknowledgment}
This work was supported in part by JST, the establishment of university fellowships towards the creation of science technology innovation, Grant Number JPMJFS2132, JSPS KAKENHI Grant Number JP22H00540, and JST ACT-X Grant Number JPMJAX22AD.

\bibliographystyle{splncs04}
\bibliography{ref}

\end{document}